\documentclass[letterpaper]{article} 
\usepackage{aaai23}  
\usepackage{times}  
\usepackage{helvet}  
\usepackage{courier}  
\usepackage[hyphens]{url}  
\usepackage{graphicx} 
\usepackage{natbib}  
\usepackage{caption} 
\usepackage{algorithm}
\usepackage{algorithmic}

\usepackage{newfloat}
\usepackage{listings}
\DeclareCaptionStyle{ruled}{labelfont=normalfont,labelsep=colon,strut=off} 
\floatstyle{ruled}
\newfloat{listing}{tb}{lst}{}
\floatname{listing}{Listing}
\title{Constructing and Interpreting \\Causal Knowledge Graphs from News}

\author {
    Fiona Anting Tan\textsuperscript{\rm 1,a},
    Debdeep Paul\textsuperscript{\rm 2,b},
    Sahim Yamaura\textsuperscript{\rm 2},
    Miura Koji\textsuperscript{\rm 2},
    See-Kiong Ng\textsuperscript{\rm 1}
}
\affiliations {
    \textsuperscript{\rm 1} Institute of Data Science, National University of Singapore\\
    \textsuperscript{\rm 2} Panasonic Industrial Devices Singapore \\
    \textsuperscript{\rm a} tan.f@u.nus.edu, \textsuperscript{\rm b} debdeep.paul@sg.panasonic.com
}

\usepackage{bibentry}

\begin{document}

\maketitle

\begin{abstract}
Many financial jobs rely on news to learn about causal events in the past and present, to make informed decisions and predictions about the future. With the ever-increasing amount of news available online, there is a need to automate the extraction of causal events from unstructured texts. In this work, we propose a methodology to construct causal knowledge graphs (KGs) from news using two steps: (1) Extraction of Causal Relations, and (2) Argument Clustering and Representation into KG. We aim to build graphs that emphasize on recall, precision and interpretability. For extraction, although many earlier works already construct causal KGs from text, most adopt rudimentary pattern-based methods. We close this gap by using the latest BERT-based extraction models alongside pattern-based ones. As a result, we achieved a high recall, while still maintaining a high precision. For clustering, we utilized a topic modelling approach to cluster our arguments, so as to increase the connectivity of our graph. As a result, instead of 15,686 disconnected subgraphs, we were able to obtain 1 connected graph that enables users to infer more causal relationships from. Our final KG effectively captures and conveys causal relationships, validated through experiments, multiple use cases and user feedback.
\end{abstract}

\section{Introduction}
\label{sec:introduction}
Many financial positions involve decision-making and prediction-making, like in stock price forecasting or sales and demand planning. Such careers require stakeholders to have a general understanding of the past, current and future market, and how one thing leads to another. Many stakeholders keep abreast of market trends by reading news. However, given the volume of the text online today, and even more if we were to consider historical news, it is impossible for any individual to consume all available information effectively. 

In the past decade, knowledge graphs (KGs) have emerged as a useful way to store and represent knowledge. By performing end-to-end causal text mining (CTM) and then representing the causal relations through a KG, it is possible to summarize the past and current events succinctly for stakeholders to learn from effectively. We define end-to-end CTM as the identification of Cause and Effect arguments in any given text, if present. 

\begin{figure*}[!h]
  \centering
  \includegraphics[width=0.95\textwidth]{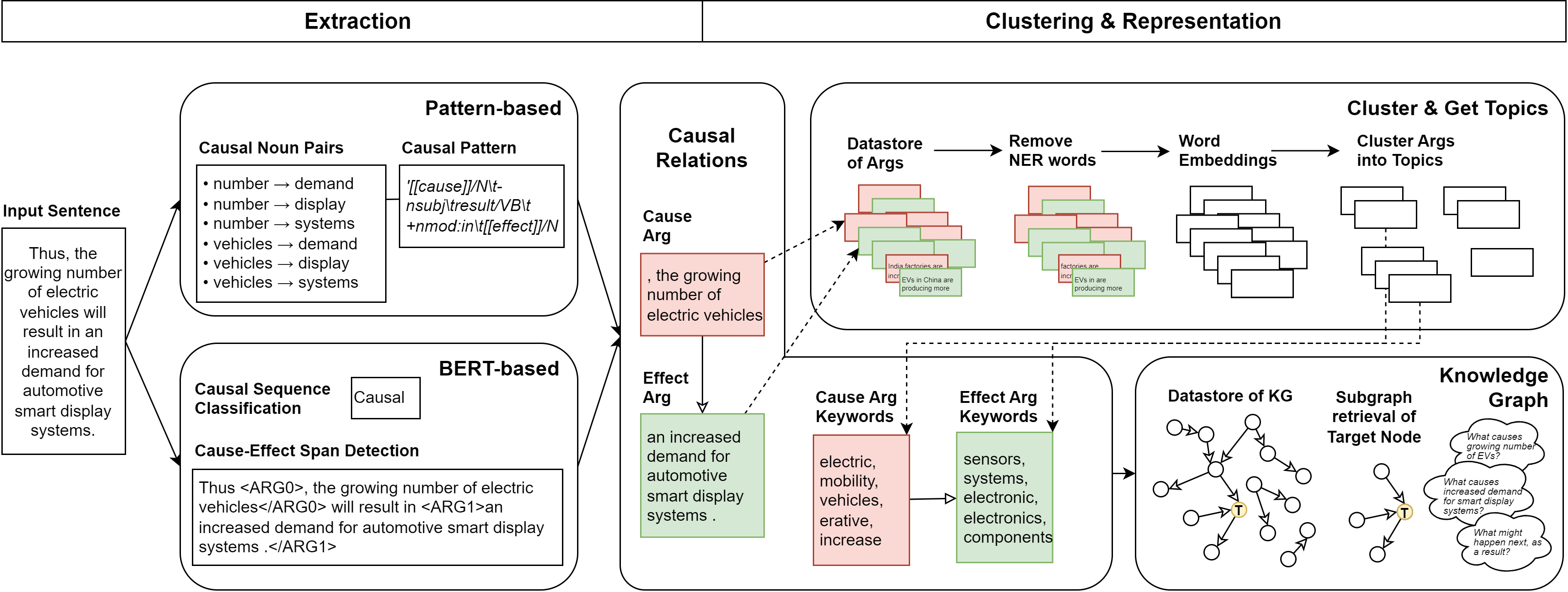}
  \caption{Overview of our methodology.}
  \label{fig:overview}
\end{figure*}

In this paper, we focus on the application of summarizing and tracking causal relations in industry news to help individuals who frequently monitor the news for market research and decision making. Therefore, the final KG constructed must be useful by being: (1) \textbf{recall-focused:} it captures a large proportion of the causal relationships present in the news, (2) \textbf{precision-focused:} the causal relationships captured are truly causal, and (3) \textbf{interpretable:} it can be used by humans to learn causal relationships. Our methodology comprises of two broad steps, shown in Figure \ref{fig:overview}: (1) Extraction of Causal Relations, and (2) Argument Clustering and Representation into Knowledge Graph. 

Our contributions are as follows:
\begin{itemize}
    \item Although many earlier works investigate construction of causal KGs from text, most utilize pattern-based methods. In our work, we employ both pattern-based and neural network-based approaches. Our findings show that the pattern-based approach drastically misses out on extracting valid causal relations compared to the neural network-based approach (1:19 ratio).
    \item Graphs built directly off extracted Cause and Effect arguments are sparse and hence, hard to interpret. To mitigate this, we investigate a simple but effective solution to cluster our arguments based on semantics to create a more connected KG that enables more causal relationships to be drawn.
    \item We evaluate our methodology on a small set of data annotated by the users, demonstrate industry use cases and discuss users' feedback on the final KG. We intend to deploy our system as a regular service to the Sales Division.
\end{itemize}

The subsequent portions of the paper are outlined as follows: We first discuss related work. Subsequently, we introduce our data and describe our methodology. Next, we present our experimental results, demonstrate multiple use cases, and finally, conclude.

\section{Related Work}
\label{sec:related}

In the recent years, many SOTA NLP solutions have been created to beat the leaderboards \cite{chen-etal-2022-ergo,zuo-etal-2020-knowdis,zuo-etal-2021-improving,zuo-etal-2021-learnda,cao-etal-2021-knowledge,chen-etal-2022-1cademy,nik-etal-2022-1cademy,aziz-etal-2022-csecu-dsg-causal}. Consistent with the general trend in NLP, the best models all use neural network architectures and pre-trained language models. Compared to pattern-based methods, neural network-based approaches can be trained to recognize more causal constructions, and therefore, in application, have a much higher recall. Yet, many papers working on constructing causal KGs still revert to rudimentary pattern-based solutions \cite{DBLP:journals/dke/IttooB13a,DBLP:conf/cikm/HeindorfSWNP20,DBLP:conf/www/RadinskyDM12,izumi-sakaji-2019-economic,xu2022data}. Recall is important in our context of monitoring news: our users have to be aware of latest causal events, and a pattern-based extraction tool with limited coverage will not be effective in identifying a high proportion of the true causal relations. In our work, we employ both pattern \cite{DBLP:conf/cikm/HeindorfSWNP20} and neural network-based \cite{unicausal} methodologies based on previous works. We investigate the differences in quality and quantity of the extracted causal relations using these two extraction approaches.

KGs can serve as a taxonomy or knowledge source to guide natural language models to make better predictions \cite{he-etal-2021-klmo-knowledge,zhang-etal-2021-eventke-event,cao-etal-2021-knowledge}. Most causal KGs are built off the extracted Cause and Effect arguments by casting them directly as nodes \cite{DBLP:conf/cikm/HeindorfSWNP20,izumi-sakaji-2019-economic,DBLP:conf/semweb/Hassanzadeh21a}. If we followed suit, we will obtain a large and poorly connected graph. In studying causality, it is beneficial to have a highly connected graph because it allows us to detect more causal relations, especially transitive ones. Additionally, generalizing over objects, actions and events allow users to make predictions of upcoming Effects even for unseen events \cite{DBLP:conf/www/RadinskyDM12}. Therefore, in our work, we condense our graphs by grouping nodes that refer to the same topic together using previous topic modelling solutions \cite{sia-etal-2020-tired,zhang-etal-2022-neural}.

\section{Dataset}
\label{sec:dataset}
We worked on 6,384 article summaries, comprising of 62,151 sentences published between 2017 and 2022. We focus on the electronics and supply-chain industry news. The articles were extracted through Google News using a webscraping tool, \texttt{Scrapy}\footnote{\url{https://docs.scrapy.org/en/latest/index.html}} on September to October 2022. We focused on the Japan, China, Europe and Global regions. The article summaries and titles were obtained using \texttt{newspaper3K}\footnote{\url{https://newspaper.readthedocs.io/en/latest/}}, which returns top 10 sentences of an article, scored and ranked using features such as sentence length, sentence position, title status, and frequency of keywords appearing in the sentence.

\section{Methodology}
\label{sec:methodology}

To briefly introduce, our approach is to extract causal relations, cluster semantically similar arguments, and store causal relations in a KG to be used for various applications. 

\subsection{Extraction of Causal Relations}
\label{ssec:extraction}

\subsubsection{Pattern-based}
We replicated CauseNet's \cite{DBLP:conf/cikm/HeindorfSWNP20} methodology of using linguistic patterns to detect causal relations. The patterns identify the shortest path between a Cause noun and an Effect noun using the dependency graph of a sentence \cite{culotta-sorensen-2004-dependency,bunescu-mooney-2005-shortest,DBLP:journals/dke/IttooB13a}. The enhanced dependency graphs were obtained using the Stanford NLP Parser \cite{chen-manning-2014-fast,schuster-manning-2016-enhanced}. The original authors extracted 53 linguistic patterns after two bootstrapping rounds on their Wikipedia dataset which we used directly.\footnote{We had to alter 28 patterns slightly to fit the our version of Stanford Parser (1.4.1 version) because \cite{DBLP:conf/cikm/HeindorfSWNP20} used an outdated 0.2.0 version.} 

To obtain more patterns, we utilized the Wikipedia dataset from \cite{DBLP:conf/cikm/HeindorfSWNP20}\footnote{\url{https://github.com/causenet-org/CIKM-20}}, which contains 1,168,155 causal sentences with the Cause and Effect arguments that were identified by the template-based method. We reverse-engineered the linguistic pattern connecting the Cause and Effect. Out of the 261,643 unique patterns we obtained, we retained the top 50 most common patterns. For the 51st to 500th pattern, we dropped patterns that have no center token (E.g. ``[[cause]]/N -nmod:of [[effect]]/N'') because such patterns that directly relate the dependency of a potential Cause to a potential Effect will return many spurious, misidentified causal relations. This reverse-engineering method provided us with 477 patterns.

Finally, the original 53 patterns were merged with the additional 477 patterns. Since 43 patterns were repeated across the two lists, the final number of patterns was 487. Most of these patterns contain causal connectives like \emph{`caused'}, \emph{`causing'}, \emph{`resulted in'} and \emph{`leading to'}. Equipped with these linguistic patterns, we extracted causal relations from news as follows:

\begin{enumerate}
    \item Extract all nouns in a sentence. We use Stanford NLP parser to obtain these part-of-speech (POS) tags.
    \item For every combination of noun pairs, identify the shortest dependency path tying the two nouns together. Format the path as a pattern string.
    \item Check if the pattern string matches with any of our causal linguistic patterns. If there is a match, the noun pair is identified to be causal.
\end{enumerate}

\begin{table}[]
\centering
\resizebox{1\columnwidth}{!}{
\begin{tabular}{p{20mm}lp{25mm}lp{25mm}p{1mm}p{25mm}}\\\hline
\multicolumn{1}{c}{Original} & & Pattern-based Extraction & & Post-processing & & Pre-processing for Clustering \\\hline
{… implementing a furlough scheme aimed at mitigating the impact of a fall in   output brought on by a global chip shortage.} & & {
$\bullet$ shortage $\rightarrow$ impact \newline
$\bullet$ shortage $\rightarrow$ fall \newline
$\bullet$ shortage $\rightarrow$ output \newline
Pattern: '[[cause]]/N -nmod:by brought/VBN +nmod:of [[effect]]'
} & & {… implementing a   furlough scheme aimed at mitigating the \texttt{<ARG1>}impact of a fall in   output\texttt{</ARG1>} brought on by a global chip   \texttt{<ARG0>}shortage \texttt{</ARG0>}.'} & & {… implementing a furlough scheme aimed at mitigating the \texttt{<ARG1>}impact of a fall in output\texttt{</ARG1>} brought \texttt{<ARG0>}on by a global chip   shortage\texttt{</ARG0>}.'}\\\hline
\end{tabular}
}
\caption{Processing of pattern-based predictions.}\label{tab:causenet_examples}
\end{table}

\emph{Post-processing} We merged arguments that have the same pattern and either Cause or Effect argument, since they refer to the same relation. To illustrate, in Table \ref{tab:causenet_examples}, the three Effect has the same Cause and the same causal pattern. Therefore, the final causal relation was processed to be \emph{``shortage''} caused \emph{``impact of a fall in output''}. Similarly, in Figure \ref{fig:overview}, the pattern-based example's six causal relations was simplified into one causal relation: \emph{``number of vehicles''} caused \emph{``demand for automotive smart display systems''}.

Pattern-based arguments tend to be short and lack the context needed for clustering. Therefore, we converted the arguments from the pattern-based approach into words from the original span up to (if occurring before) or until (if occurring after) the signal words. For example, in Table \ref{tab:causenet_examples}, the Cause argument was altered to start right before the signal word \emph{`brought'}. Since the Effect argument already spans up to the signal word, it remains the same.

To conclude this subsection, the pattern-based matching approach allowed us to identify 1,006 sentences and 975 causal relations from 611 unique sentences.

\subsubsection{BERT-based}

UniCausal \cite{unicausal}\footnote{\url{https://github.com/tanfiona/UniCausal}} is a causal text mining repository that consolidated six datasets (AltLex \cite{hidey-mckeown-2016-identifying}, BECAUSE 2.0 \cite{dunietz-etal-2017-corpus}, CausalTimeBank \cite{mirza-etal-2014-annotating,mirza-tonelli-2014-analysis}, EventStoryLine \cite{caselli-vossen-2017-event}, Penn Discourse Treebank V3.0 \cite{webber2019penn} and SemEval2010Task8 \cite{hendrickx-etal-2010-semeval}) for three tasks (Causal Sentence Classification, Causal Pair Classification, and Cause-Effect Span Detection). Pre-trained models were created and made available online. These models were trained on each task independently, builds on BERT-based pre-trained encoders \cite{devlin-etal-2019-bert}, and used Cross Entropy Loss. All six datasets were used for training and testing. In our work, we utilized three pre-trained models developed by UniCausal:
\begin{enumerate}
    \item \textbf{Causal Sentence Classification (CSC):} Model identifies if a sentence contains causal relations or not. After passing the sentence through BERT-encoder layers, the embeddings of the \texttt{[CLS]} token are processed through a dropout layer, followed by a classification layer to generate predicted logits. The pre-trained model reported 70.10\% Binary F1 score.
    \item \textbf{Causal Pair Classification (CPC):} Model identifies if a pair of arguments (\texttt{ARG0}, \texttt{ARG1}) that are marked in the sentence are causally related or not, such that \texttt{ARG0} causes \texttt{ARG1}. It follows the same architecture as CSC. The pre-trained model reported 84.68\% Binary F1 score.
    \item \textbf{Cause-Effect Span Detection (CESD):} Model identifies the consecutive span of words that refer to the Cause and Effect arguments. Framed as a token classification task, after the BERT-encoder layers, the sequence output is fed through a dropout then classification layer to obtain the predicted logits per token. The pre-trained model reported 52.42\% Macro F1 score.
\end{enumerate}

To extract causal relations from text, we applied both CSC and CESD predictors to all sentences. For causal sentences identified by CSC, we retained the cause and effect arguments identified by CESD.

\emph{Post-processing} One limitation of UniCausal's CESD is that it was designed to predict only one causal relation per example. However, in our investigations, many instances had multiple \texttt{ARG0} and \texttt{ARG1} predictions. Without additional information, the relationship between the multiple causes and effects was unclear. Therefore,  we implemented a post-processing procedure involving three steps: (1) Merge sequential arguments, (2) Keep longest argument for examples with three arguments, and (3) Keep multiple causal relations based on CPC predictions. Additionally, we utilized CPC to identify and retain additional causal examples. Details about these procedures can be found in the Appendix.

Altogether, the BERT-based method identified 19,250 sentences with 19,192 causal relations from 15,702 unique sentences.



\subsection{Argument Clustering}
\label{ssec:clustering}
We wish to cluster the arguments that have similar meaning, both in terms of the topic mentioned in the argument (E.g. supply, profits, automotives, etc.) and the impact on it (E.g. positive, negative, etc.). We used the approach by \cite{sia-etal-2020-tired,zhang-etal-2022-neural} to generate word embeddings from sequences and cluster the embeddings directly.

\paragraph{Neutralizing named-entities}
We are not interested to cluster arguments that refer to the same organization, location, or date. Thus, we used the 7-class Stanford Named Entity Recognition (NER) Tagger \cite{finkel-etal-2005-incorporating} \footnote{\url{https://nlp.stanford.edu/software/CRF-NER.shtml}} to extract named-entities for locations, persons, organizations, times, money, percents, and dates. Subsequently, we remove the words corresponding to any of these entities in the argument spans. For example, the Cause argument \emph{``to jointly produce premium EVs in China''} was converted to \emph{``to jointly produce premium EVs in''}. Note that in the final KG, the original arguments were used.

\paragraph{Word embeddings} To generate word embeddings that clusters semantically similar arguments together and semantically different arguments apart, we used the supervised pre-trained language model by SimCSE \cite{gao-etal-2021-simcse} to encode our NER-neutralized arguments into embeddings. SimCSE was trained to identify whether the relationship betweent two sentences suggests entailment, neutral, or contradition. SimCSE was evaluated against standard semantic textual similarity tasks, and achieved an average 81.6\% Spearman’s correlation, a 2.2\% improvement compared to previous best results. Our embeddings had a feature dimension of 786 because the model is built on the BERT model, \texttt{bert-base-uncased}. 

\paragraph{Clustering and getting keywords per cluster}
Similar to \cite{sia-etal-2020-tired}, we used K-Means to cluster the 35,230 embeddings into 3,000 topics. We remove relations where the Cause and Effect fall under the same topic so that we do not have nodes that connected to itself. To obtain the top keywords per topic, we used the TFIDF $\times$ IDF method proposed by \cite{zhang-etal-2022-neural}:

\begin{equation}
    TFIDF_d = \frac{n_w}{\sum_{w} n_w} \cdot log(\frac{|D|}{|\{ d \in D | w \in d \}|})
\end{equation}
\begin{equation}
    IDF_k = log(\frac{|K|}{|\{ w \in K \}|})
\end{equation}
\begin{equation}
    TFIDF \times IDF = TFIDF_d \cdot IDF_k
\end{equation} where $n_w$ is the frequency of word $w$ in each document $d$, $D$ refers to the whole corpus, $|K|$ is the number of clusters, and $|\{ w \in K \}|$ is the number of clusters where word $w$ appears in. This approach helps to identify the important words to each cluster compared to the rest of the dataset ($TFIDF_d$), while penalizing frequent words that appear in multiple clusters ($IDF_k$). Empirical findings demonstrate this method significantly outperforms regular TF or TFIDF methods in selecting topic words \cite{zhang-etal-2022-neural}. In the end, we extract 5 keywords per topic, which corresponds to the text displayed in the nodes of a graph. If the cluster contains only one argument, then the argument text is displayed instead.

\subsection{Knowledge Graph}
\label{ssec:graph_creation}
We define our knowledge graph $G = (V,E)$ as a collection of nodes $V = \{(v_1, v_2, ..., v_n)\} $ and directed edges $E = \{(v_1,v_2), (v_2, v_3), ...\}$. A directed edge $(v_i, v_j)$ represents the presence of causality between the two nodes, where $v_i$ is the Cause and $v_j$ is the Effect. The edges are also weighted by $s$, which represents the number of sentences in our dataset that has been identified to convey that $v_i$ causes $v_j$. 

Table \ref{tab:statistics} shows the statistics of our extracted relations and constructed KG. Earlier, out of 62,151 sentences, we identified 15,902 unique sentences containing 20,086 causal relations. Before clustering, a KG built directly on these extracted relations would have 35,230 unique nodes and 20,086 unique edges, with an average support per edge of 1.008. By performing argument clustering, we created a highly connected KG, with 3,000 unique nodes, 17,801 unique edges, and an average support per edge of 1.122. Table \ref{tab:statistics_kg} displays some graph statistics before and after clustering. Again, we observe that the KG after clustering is denser and more connected. In fact, instead of 15,686 subgraphs, our KG is now represented by 1 connected graph. Visualizations of the KGs in the later sections are performed using Cytoscape\footnote{\url{https://cytoscape.org/}}, an open-source software for visualizing and interacting with graphs.

\begin{table*}[]
\centering
\resizebox{2\columnwidth}{!}{
\begin{tabular}{p{22mm}|lp{10mm}p{10mm}p{10mm}p{13mm}|lp{10mm}p{10mm}p{13mm}|lp{10mm}p{10mm}p{13mm}}\hline
\multicolumn{6}{c|}{Extraction} &  & \multicolumn{3}{c|}{Before Clustering} &  & \multicolumn{3}{c}{After Clustering} \\\hline
\multicolumn{1}{c|}{Method} & \multicolumn{1}{c}{} & \multicolumn{1}{c}{$|Sents|$} & \multicolumn{1}{c}{$|Sents|$} & \multicolumn{1}{c}{$|Rels|$} & \multicolumn{1}{p{13mm}|}{Avg Rel Support} & \multicolumn{1}{c}{} & \multicolumn{1}{c}{$|V|$} & \multicolumn{1}{c}{$|E|$} & \multicolumn{1}{p{13mm}|}{Avg E Support} & \multicolumn{1}{c}{} & \multicolumn{1}{c}{$|V|$} & \multicolumn{1}{c}{$|E|$} & \multicolumn{1}{p{13mm}}{Avg E Support} \\\hline
Pattern-based & & 1,006 & 611 & 975 & 1.032 &  & 1,476 & 975 & 1.032 &  & 774 & 845 & 1.340 \\
BERT-based &  & 19,250 & 15,702 & 19,192 & 1.003 &  & 33,940 & 19,192 & 1.003 &  & 2,990 & 17,075 & 1.120 \\
Total &  & 20,255 & 15,902 & 20,086 & 1.008 &  & 35,230 & 20,086 & 1.008 &  & 3,000 & 17,801 & 1.122\\\hline
\end{tabular}
}
\caption{Summary statistics of extracted causal relations per step.}\label{tab:statistics}
\end{table*}
\begin{table}[]
\centering
 \resizebox{1\columnwidth}{!}{
\begin{tabular}{l|c|c}\hline
& Before Clustering & After Clustering \\\hline
No. of Unique Nodes, $|V|$                      & 35,230             & 3,000             \\
No. of Unique Edges, $|E|$                      & 20,086             & 17,801            \\
Total   Weight, $\sum s$               & 20,254             & 19,965            \\
No. of   Subgraphs           & 15,686             & 1                \\
Avg   Clustering Coefficient & $9.81e^{-06}$          & $1.75e^{-02}$         \\
Avg Degree   Centrality      & $3.24e^{-05}$          & $3.96e^{-03}$         \\
Avg   Eigenvector Centrality & $6.64e^{-05}$          & $1.32e^{-02}$         \\
Transitivity                 & $4.17e^{-04}$          & $8.81e^{-03}$       \\\hline
\end{tabular}
}
\caption{Graph statistics before and after clustering.}\label{tab:statistics_kg}
\end{table}


\section{Experimental Results}
\label{sec:results}

\subsection{Extraction of Causal Relations}
\label{ssec:results_extraction}
\paragraph{Quantitative evaluation}
We asked users to randomly select 15 articles from the Google News dataset and annotate the Causes and Effects per sentence. 49 causal relations from 43 sentences were identified. 6 sentences had more than one causal relation. A correct case is one where the model and human annotations have $\geq 1$ word(s) overlapping for both Cause and Effect spans. We could then calculate the number of True Positives (TP), False Positives (FP) and False Negatives (FN) by treating the human annotations as the gold standard. True Negative (TN) is 0 in all cases because we do not evaluate on non-causal sentences and relations. Appendix Table \ref{tab:extraction_eval_examples} provides some examples comparing model predictions to human annotations. Subsequently, we calculate Precision (P), Recall (R), and F1 scores using the following formulas:


\begin{equation}
    P = \frac{TP}{TP+FP},\hspace{2.5mm} R = \frac{TP}{TP+FN},\hspace{2.5mm} F1 = \frac{2 \times P \times R}{P+R}
\end{equation}




\begin{table}[]
\centering
\resizebox{0.9\columnwidth}{!}{
\begin{tabular}{p{32mm}p{10mm}p{10mm}p{10mm}}\hline
Extraction Method & P & R & F1 \\\hline
Pattern-based & \textbf{100.00} & 4.08 & 7.84 \\
BERT-based & 76.09 & 71.43 & 73.68 \\\hline
Both & 75.00 & \textbf{73.47} & \textbf{74.23}\\\hline
\end{tabular}
 }
\caption{Performance metrics for extraction on human-annotated test set. Scores are reported in percentages (\%). Top score per column is bolded.}\label{tab:extraction_results}
\end{table}

Our model identified 48 causal relations from 43 sentences. 4 sentences had more than one causal relation. The performance metrics are reported in Table \ref{tab:extraction_results}. By using our proposed method of combining both pattern-based and BERT-based approaches in extraction, the F1 score is the highest at 74.23\%.

Although the pattern-based has very good precision (100\%), it could only identify 2/49 causal relations, resulting in an extremely low recall score of 4\%. In the application of monitoring news and trends, such a low recall is unacceptable as it would miss out on many key happenings. The BERT-based approach extracted much more causal relations than the pattern-based method, scoring a higher recall of 71.43\%. This is because UniCausal is trained on a large dataset, and its architecture also allows models to learn from varied linguistic structures, including implicit causal relations. Consistent with our findings, we observe in Table \ref{tab:statistics} that the BERT-based approach extracted 19x more causal relations than the pattern-based approach for our whole dataset.

The model proposed 12 causal relations that were not annotated by the humans. Upon checking, 5/11 were correct in that the Cause and Effect spans suggested are causal. However, they are duplicates arising from the post-processing done for BERT-based extraction that accepts any pair of arguments that CPC detects as causal. Therefore, when comparing against the human annotated test set, these duplicates were treated as spurious relations. If we consider these five examples as Correct, then the overall precision would increase to 85.42\%.





\subsection{Argument Clustering}
\label{ssec:results_clustering}
\paragraph{Quantitative evaluation}
From the 36 causal relations where the model and users were in agreement with, we asked users to group the arguments with similar meaning and give a topic label to each group. The users clustered 72 arguments into 50 topics. Some example topics are: \emph{`increased competition'}, \emph{`taxation'}, \emph{`cost reduction'}, \emph{`business hurdles'} and \emph{`raw material shortage'}. To obtain the model's predictions, we filtered out the nodes of the 36 causal relations from the whole KG (described in the "Knowledge Graph" Section). The 72 arguments were clustered into 70 topics. We compare the model's and user's clustering using Normalized Mutual Information (NMI), an entropy-based evaluation metric. Because most clusters only contain one span from both the model's and user's clustering, NMI is high at 93.62\%. However, given the small sample size, this score can be misleading. More annotated data is needed to evaluate clustering performance.

\paragraph{Qualitative evaluation}
Due to limited space, we summarize our qualitative experiments and findings in this subsection. Details are available in the Appendix.

In Table \ref{tab:statistics_kg}, we show that clustering helps to increase the average edge weight and node centrality. Our argument clustering solution helps to create a highly connected causal KG, which is more insightful to infer causal relationships from. For example, before clustering, we could only infer that \emph{`pandemic'} causes \emph{`disruptions'}. After clustering, our subgraph detected that \emph{`pandemic'} causes supply chain disruptions, chip and semiconductor shortages, sales decreases, and general interferences and disruptions.

We also found that argument clusters are much more defined on a 2D plot after we remove the named-entities from arguments. Named-entity removal allows the clustering process to focus on more meaningful words referring to the event, sentiment, or topic instead.

\section{Applications in the Industry}
\label{sec:discussion}

\subsection{Use Cases}
\label{ssec:results_usecases}
\paragraph{Summarization} 
Our causal KG is useful for summarizing reported causal relations in news. In the earlier section about qualitative evaluation, we constructed a \emph{`pandemic'} subgraph and demonstrated how we can swiftly learn about the reported effects of pandemic.

\paragraph{Answering causal questions and predicting future events} 
Our KG is also useful for answering causal questions. In Figure \ref{fig:overview}, users that learn that that Event A (\emph{``the growing number of electric vehicles''}) causes Event B (\emph{``an increased demand for automotive smart display systems''}) might ask: What might happen next as a result of Event B? By setting the target node to be Event B, we identify that the next two likely events (based on edge support) are that display systems will \emph{``become an integral part of the automotive supply chain''} (\emph{`automotive\_industry\_shaft\_xa\_siness'}) and that this \emph{``trend is expected to continue during the forecast period''} (\emph{`forecast\_period\_during\_anticipated\_analysts'}). Other subsequent Effects are: \emph{``so does demand for wiring harnesses and related electronic sub-assemblies''}, \emph{``forcing European car makers to rely on Asian suppliers''}, \emph{``the task of designing today’s cars much more difficult''}, and many more meaningful predictions. We can also ask other causal questions like ``Are there other causes of Event B?'' and ``What caused Event A in the first place?''

\paragraph{Inferring transitive causal relations}
In transitive relations, if Event A causes Event B, and Event B causes Event C, then Event A can also be said to cause Event C. For causal relations, the transitive property fails if the relations are too specific\footnote{Example of specific causal relations violating transitivity: \emph{``Sugar makes John happy. Sugar causes diabetes. Diabetes makes John sad. Does sugar make John happy or sad?''}}. Referring to the example from the earlier paragraph, we observe that transitivity does hold: \emph{``the growing number of electric vehicles''} (Event A) does help make automotive smart display systems \emph{``become an integral part of the automotive supply chain''} (Event C) through \emph{``an increased demand for automotive smart display systems''} (Event B). This observation has strong implications on how insightful our KG can be for inferring causal relationships that were not otherwise stated directly in the news. Further analysis will be needed to identify the cases where transitivity fails, and how we should handle them.

\begin{figure}[!h]
  \centering  \includegraphics[width=1\columnwidth]{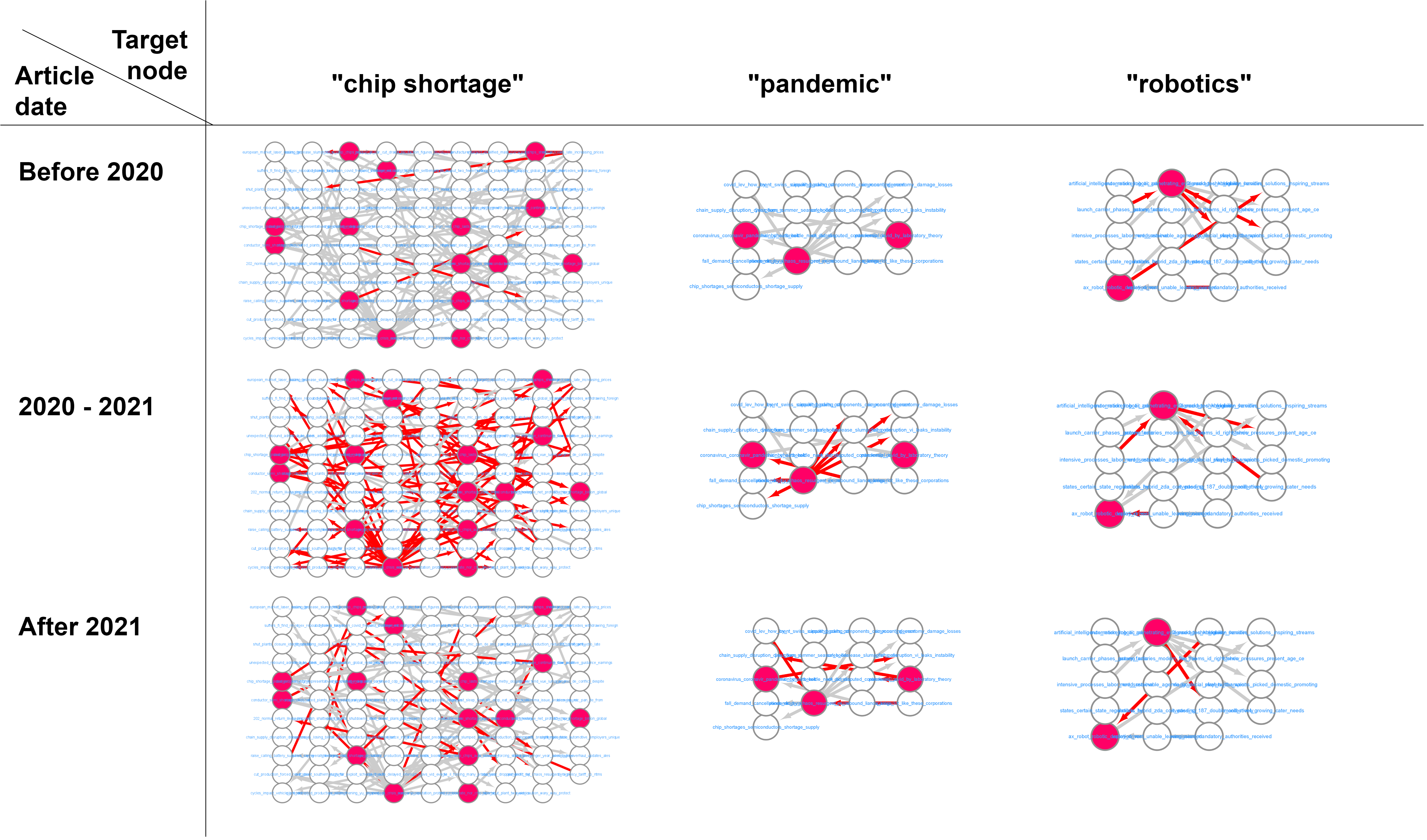}
  \caption{Subgraph(s) filtered based on nodes that are one step away from target node(s) highlighted in pink. In this example, a node is referred to as a target if it contains the search term in any order indicated on top (E.g. \emph{`chip shortage'}). Edges from articles falling under the time period corresponding to the right axis are highlighted in red.}
  \label{fig:trends_monitoring}
\end{figure}

\paragraph{Trend monitoring}
We demonstrate the potential of our KG to reflect trends over time. To conduct the experiment, we split our dataset into articles that are published across three time baskets: Before 2020, 2020 to 2021 (inclusive), and after 2021. Since our dataset is very imbalanced across time, we down-sampled the two larger baskets such that all baskets have the same sample size. Similar to analyses before, we created subgraphs by filtering out target nodes and nodes that are one step away from target node(s). A node is a target if the search term(s) can be found in the node description in any order. In Figure \ref{fig:trends_monitoring}, we studied three search terms across the three time baskets. For each subgraph, the edges highlighted in red falls within the time basket of interest. Our findings show that the frequency of causal relationships about the topic \emph{``chip shortage''} was rare before 2020, extremely heated during the pandemic period of 2020 to 2021, and lower from 2021 onwards. This is validated by experts' understanding that the COVID-19 pandemic kick-started the chip shortage, amongst many other reasons. As a sanity check, we found that no causal relationships were mentioned about \emph{``pandemic''} before 2020. This makes sense because the awareness of COVID-19 only started taking off in the first quarter of 2020. As a control, we also checked that interest in topics about \emph{``robotics''} stayed roughly constant throughout the three time baskets. To conclude, our KG can be helpful for monitoring heated causal topics and news trends across time.

\subsection{User Feedback}
\label{ssec:results_feedback}
The final KG was presented users to gather feedback. Response was positive, and many users could find a use for this KG in their daily work. We intend to deploy the our system to generate regular snapshots of the news that will serve as market reports for the Sales Division. Users believe that harnessed with knowledge about recent causal events, they can improve their market understanding and perform better at prediction-related tasks, like sales and demand forecasting. In the future, users would also like to see the KG to be improved by adding temporal and sentiment elements.

\section{Conclusion \& Future Work}
\label{sec:conclusion}
We focused on the application of extracting causal relations in industry news to construct a causal KG. Our approach was (1) recall-focused, by employing BERT-based on top of pattern-based extraction methods. Our approach was also (2) precision-focused, and for our test set, achieved 75\% score. Finally, our final KG was designed to be (3) interpretable, with many use cases and was deemed useful by our users in the industry. 

Our work can be replicated onto many other domains. In the future, we intend to annotate a larger test set for more concrete evaluation. Additionally, we plan to deploy the extraction and graphing system to generate monthly snapshots of the market. We will monitor the users' interactions with the system and identify areas for improvement. Lastly, we hope to include temporal and sentiment elements in our extraction to enrich the final KG.

\bibliography{bibs/anthology, bibs/custom}

\newpage

\appendix

\section{Appendix}

\subsection{Post-processing of BERT-based predictions}
\label{apx:post-process-bert}

\begin{table*}[]
\centering
\resizebox{2\columnwidth}{!}{
\begin{tabular}{p{3mm}p{25mm}lp{3mm}p{40mm}lp{60mm}lp{15mm}}\\\hline
\multicolumn{1}{c}{S/N} & \multicolumn{1}{c}{Original} &  & \multicolumn{2}{c}{BERT-Based Extraction} &  & \multicolumn{3}{c}{Post-Processing} \\\cline{4-5}\cline{7-9}
 &  &  & \multicolumn{1}{c}{CSC} & \multicolumn{1}{c}{CESD} &  & \multicolumn{1}{c}{Final} &  & \multicolumn{1}{c}{Method} \\\hline
1 & {
Still, car companies and dealers may have to eventually adopt some of the changes Tesla has introduced to win over buyers who have grown used to buying cars online.
} &  & 1 & {
\texttt{<ARG1>}Still\texttt{</ARG1> }, car companies and dealers may have to \texttt{<ARG1>}eventually adopt some of the changes Tesla has introduced\texttt{</ARG1>} \texttt{<ARG0>}to win over buyers who have grown used to buying cars online .\texttt{</ARG0>}
} &  & {
\texttt{<ARG1>}Still , car companies and dealers may have to eventually adopt some of the changes Tesla has introduced\texttt{</ARG1>} \texttt{<ARG0>}to win over buyers who have grown used to buying cars online .\texttt{</ARG0>}
} &  & Merge Sequential   Arguments \\\hline
2 & ``Due to the current situation in this region, there may be disruptions in the supply chain.'' &  & 1 & \texttt{<ARG1>}``\texttt{</ARG1>}   Due to \texttt{<ARG0>}the current situation in this region\texttt{</ARG0>}   \texttt{<ARG1>}, there may be disruptions in the supply chain .   ''\texttt{</ARG1>} &  & `` Due to   \texttt{<ARG0>}the current situation in this region\texttt{</ARG0>} \texttt{<ARG1>}, there may be disruptions in the supply chain . ''\texttt{</ARG1>} &  & Keep Longest Argument   for 3-Argument Examples \\\hline
3 & Hence we expect global supply chains switching to EVs to have a positive impact on the Indian EV industry. &  & 1 & Hence we expect   \texttt{<ARG0>}global\texttt{</ARG0>} \texttt{<ARG1>}supply\texttt{</ARG1>} \texttt{<ARG0>}chains switching to EVs to\texttt{</ARG0>} \texttt{<ARG1>}have   a\texttt{</ARG1>} \texttt{<ARG0>}positive impact on\texttt{</ARG0>} \texttt{<ARG1>}the  Indian EV industry\texttt{</ARG1>} \texttt{<ARG0>}.\texttt{</ARG0>} &  & Hence we expect   \texttt{<ARG0>}global supply chains switching to EVs\texttt{</ARG0>} to have a   positive impact on \texttt{<ARG1>}the Indian EV industry\texttt{</ARG1>}. &  & Keep Multiple Causal Relations based on CPC \\\hline
4 & China stocks rose on Monday after the governor of the country’s central bank vowed to increase the implementation of prudent monetary policy to support the real economy. &  & 1 & \texttt{<ARG1>}China\texttt{</ARG1>}   stocks rose on Monday after \texttt{<ARG0>}the\texttt{</ARG0>} governor of the   \texttt{<ARG0>}country\texttt{</ARG0>} ’ s central bank vowed \texttt{<ARG1>}to increase the implementation of prudent monetary policy\texttt{</ARG1>}   \texttt{<ARG0>}to support the real economy .\texttt{</ARG0>} &  & {
$\bullet$ China stocks rose on   Monday after \texttt{<ARG0>}the governor of the country’s\texttt{</ARG0>} central  bank vowed \texttt{<ARG1>}to increase the implementation of prudent monetary   policy\texttt{</ARG1>} to support the real economy.\newline
$\bullet$ China stocks rose on Monday after the governor of the country’s central   bank vowed \texttt{<ARG1>}to increase the implementation of prudent monetary   policy\texttt{</ARG1>} \texttt{<ARG0>}to support the real economy. \texttt{</ARG0>}\newline
$\bullet$ \texttt{<ARG1>}China stocks rose on Monday after the governor of the country’s   central bank vowed to increase the implementation of prudent monetary   policy\texttt{</ARG1>} \texttt{<ARG0>}to support the real economy. \texttt{</ARG0>}
} &  & Keep Multiple Causal   Relations based on CPC \\\hline
5 & That break was   extended as the virus spread. &  & 0 & \texttt{<ARG1>}That   break was extended\texttt{</ARG1>} as \texttt{<ARG0>}the virus spread.\texttt{<ARG0>} &  & \texttt{<ARG1>}That   break was extended\texttt{</ARG1>} as \texttt{<ARG0>}the virus spread.\texttt{<ARG0>} &  & Add Causal Relations   based on CPC \\\hline
6 & Ford is shutting its   car factories in India after Ford India racked up more than \$2bn in losses   over the past decade. &  & 0 & \texttt{<ARG1>}Ford is   shutting its car factories in India\texttt{</ARG1>} after \texttt{<ARG0>}Ford India   racked up more than \$2bn in losses over the past decade.\texttt{</ARG0>} &  & \texttt{<ARG1>}Ford is   shutting its car factories in India\texttt{</ARG1>} after \texttt{<ARG0>}Ford India   racked up more than \$2bn in losses over the past decade.\texttt{</ARG0>} &  & Add Causal Relations   based on CPC\\\hline
\end{tabular}
}
\caption{Processing of BERT-based predictions.}\label{tab:unicausal_examples}
\end{table*}

One limitation of UniCausal's CESD is that it was trained to predict only one causal relation per example. The predictions are easy to infer if only one \texttt{ARG0} and one \texttt{ARG1} is predicted by the model. However, in practice, there were no restrictions on the number of arguments that can be predicted. In our investigations, many examples had multiple \texttt{ARG0} and multiple \texttt{ARG1} predicted. Without further information on which argument is tied to which, we cannot identify the Cause and the Effect.

Therefore, for examples with multiple \texttt{ARG0} and/or multiple \texttt{ARG1}, we had to perform heuristics to process the predictions into usable Cause and Effect arguments. We employed three key post-processing steps, with examples shown in Table \ref{tab:unicausal_examples}, and explained below:

\begin{enumerate}
    \item Merge Sequential Arguments: Based on sequence in text, if two arguments follow one another and are of the same type, we merge them into one argument. See Example 1.
    \item Keep Longest Argument for 3-Argument Examples: For examples where there are (A) Two \texttt{ARG0} and one \texttt{ARG1} or (B) One \texttt{ARG0} and two \texttt{ARG1}, we kept the longest argument (in terms of character count) for the type that had two arguments. See Example 2.
    \item Keep Multiple Causal Relations based on CPC: For examples with multiple arguments, we retained all \texttt{<ARG0>} as potential Causes and all \texttt{<ARG1>} as potential Effects. We also retained two disconnected potential Causes with an Effect in the middle as a potential Cause. Likewise, we retained two disconnected potential Effects with a Cause in the middle as a potential Effect. With a list of potential Causes and Effects, we marked the original sentence with the potential Cause and Effect, and fed the marked sentence through a CPC model to predict if the pairs are causal or not. We retained pairs that were predicted to be causal. See Examples 3 and 4. In the later sections, we refer to all causal relations extracted from this method as \texttt{BERT-M}.
\end{enumerate}


During investigations, we also realised by filtering away examples that CSC predicts as non-causal early on, we are losing out on potential causal examples. To rectify this, we retained examples where CESD identified one Cause and Effect, suggesting the Cause/Effect boundaries were easy to identify and more likely to represent a causal event. After marking the original sentence with these potential arguments, we fed the marked sentence through the CPC predictor to obtain a prediction of whether the pair of arguments are causal or not. We retained pairs that were predicted to be causal. See Examples 5 and 6. 

\subsection{Quantitative evaluation of causal relation extraction}

Table \ref{tab:extraction_eval_examples} provides some examples comparing model predictions to human annotations, and how the example contributes to the final score in terms of TP, FP and FN. 

\begin{table*}[]
\centering
\resizebox{1.95\columnwidth}{!}{
\begin{tabular}{p{5mm}p{68mm}p{68mm}p{8mm}}\hline
S/N & Human   Annotations & Model Predictions & Score \\\hline
1 & \texttt{<ARG1>}Toyota in India   has largely pivoted toward hybrid vehicles, which attract taxes of as much as   43\%\texttt{</ARG1>} because \texttt{<ARG0>}they aren’t purely   electric\texttt{</ARG0>}. & Toyota in India has largely pivoted toward hybrid   vehicles\texttt{<ARG1>}, which attract taxes of as much as 43\%   \texttt{</ARG1>}because \texttt{<ARG0>}they aren’t purely electric.\texttt{</ARG0>} & 1 TP \\\hline
2 & \texttt{<ARG0>}Wednesday’s vote   will determine\texttt{</ARG0>} \texttt{<ARG1>}how fast companies will have to   transition production and when they will have to stop selling combustion engine vehicles\texttt{</ARG1>}. & \emph{None} & 1 FN \\\hline
3 & {$\bullet$ \texttt{<ARG0>}The French furlough deal, agreed by four unions at Stellantis\texttt{</ARG0>}, \texttt{<ARG1>}enables the company to reduce the number of hours worked by staff   affected by the chip shortage\texttt{</ARG1>}. \newline
$\bullet$ The French furlough deal,   agreed by four unions at Stellantis, enables the company to reduce \texttt{<ARG1>}the number of hours worked by staff\texttt{</ARG1>} \texttt{<ARG0>}affected by the chip shortage\texttt{</ARG0>}.
} & {
$\bullet$  \texttt{<ARG0>}The French furlough deal, \texttt{</ARG0>}agreed by four unions at   Stellantis, enables \texttt{<ARG1>}the company to reduce the number of hours   worked by staff affected by the chip shortage.\texttt{</ARG1>} \newline
$\bullet$ The French   furlough deal, agreed by four unions at Stellantis, enables \texttt{<ARG1>}the   company to reduce the number of hours worked by staff affected by the \texttt{</ARG1>}\texttt{<ARG0>}chip shortage\texttt{</ARG0>}.\newline
$\bullet$ \texttt{<ARG0>}The French furlough deal, \texttt{</ARG0>}agreed by   four unions at Stellantis, enables \texttt{<ARG1>}the company to reduce the   number of hours worked by staff affected by the \texttt{</ARG1>}chip shortage.
} & 2 TP, 1 FP \\\hline
4 & \emph{None} & The ambition for its Circular Economy Business Unit is   \texttt{<ARG0>}to quadruple extended life revenues for parts and services   \texttt{</ARG0>}\texttt{<ARG1>}and multiply recycling revenues tenfold by 2030 as   compared to 2021.\texttt{</ARG1>} & 1 FP\\\hline
\end{tabular}
 }
\caption{Examples of human annotations compared to model predictions, and how they contribute to the True Positive (TP), False Positive (FP) and False Negative (FN) counts.}\label{tab:extraction_eval_examples}
\end{table*}

\subsection{Qualitative evaluation of causal relation extraction}
\label{apx:qual-eval}
In this Section, we outline the qualitative experiments and findings to assess the effectiveness of our argument clustering approach. This Section is summarized briefly under the Section "Experimental Results -- Argument Clustering -- Qualitative evaluation".

\begin{figure*}[!h]
  \centering
  \includegraphics[width=1.9\columnwidth]{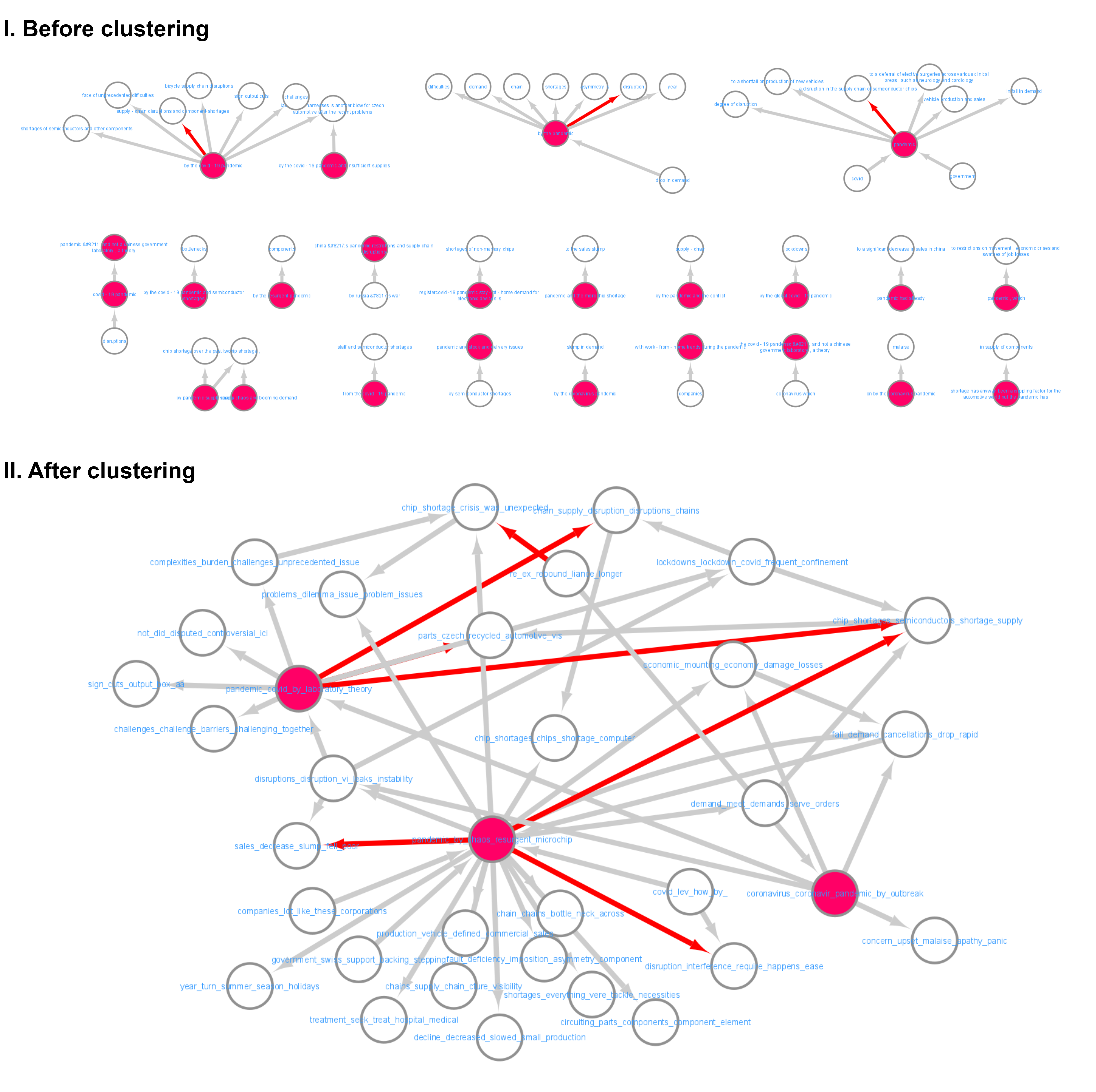}
  \caption{Subgraph(s) filtered based on nodes that are one step away from target node(s) highlighted in pink. In this example, a node is referred to as a target if it contains the keyword \emph{`pandemic'}. Edges with support of $\geq 2$ are highlighted in red.}
  \label{fig:clustering_subgraphs}
\end{figure*}

\paragraph{\emph{Effect of clustering on KG interpretability}}
In Table \ref{tab:statistics_kg}, we show that clustering helps to increase the average edge weight and node centrality. The resulting effect is that we obtained a single connected graph instead of multiple subgraphs. We visualize this phenomenon in this subsection, and demonstrate why a highly connected graph is more useful for inferring causal relationships. Figure \ref{fig:clustering_subgraphs} compares subgraphs before and after clustering. We set target nodes to be any node that contains the keyword \emph{`pandemic'}. Subgraphs are then defined as target nodes plus nodes that are one step away from target nodes. Before clustering, we obtain multiple disconnected subgraphs that contain either a Cause or Effect related to the keyword \emph{`pandemic'}. After clustering, we end up with only three target nodes that reflect pandemic related traits, reflected in one highly connected graph.

Before clustering, the three edges that had a support of $\geq 2$ all conveyed the idea that \emph{`pandemic'} causes \emph{`disruptions'}. However, after clustering, we could obtain much more meaningful causal relations. Our subgraph detected that \emph{`pandemic'} causes supply chain disruptions (\emph{`chain\_supply\_disruption\_disruption\_ chains'}), chip and semiconductor shortages (\emph{`chip\_shortages\_semiconductors\_ shortage\_supply'}), sales decreases (\emph{`sales\_decrease\_slump\_fell\_poor'}), and disruptions and interferences in general (\emph{`disruption\_interference\_require\_happens \_ease'}). Therefore, our argument clustering solution helps to create a highly connected causal KG, which is more insightful to infer causal relationships from.

\begin{figure*}[!h]
  \centering
  \includegraphics[width=1.5\columnwidth]{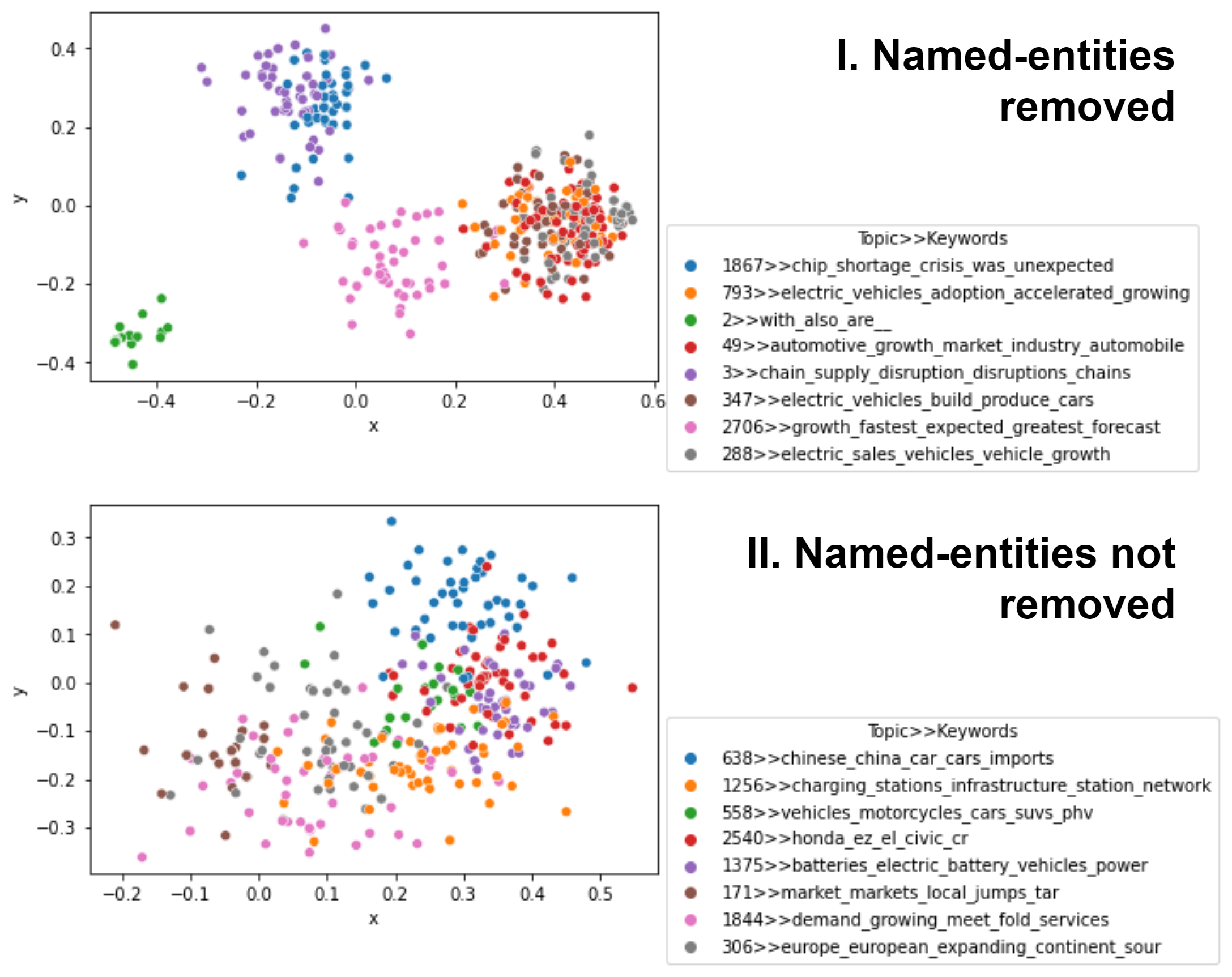}
  \caption{Scatterplot of arguments from top 8 nodes (in terms of number of connected edges) cast on the first two components using Principal Component Analysis of the word embeddings. Each point reflects an argument, and is colored by the node/topic they belong to.}
  \label{fig:clustering_pca}
\end{figure*}

\paragraph{\emph{Effect of named-entity removal on clustering}}
Figure \ref{fig:clustering_pca} compares the clustering outcomes with and without the removal of named-entities. Compared to Panel II, Panel I has well-defined clusters. In Panel I, Topics 3 (Purple) and 1867 (Blue) overlap closely because supply chain disruptions and shortages often refer to similar contexts related to manufacturing and production. Topics 49 (Red), 347 (Brown) and 793 (Orange) refers to topics about the automotive and electronic vehicles (EVs) industry, and hence, occur closely. Interestingly, Topic 2 (Green) reflects arguments that are short and make little sense (E.g. \emph{``t sm c''}), or is wholly comprised of named-entities and thus becomes an empty span (E.g. \emph{``Tesla''}, \emph{``Europe''}, \emph{``Aug. 17''}, \emph{``November''}). Thus, the points under Topic 2 are far from all other points in the scatterplot. In Panel II, the clusters are less defined. The topics' keywords include named-entities like \emph{``China''}, \emph{``Honda''} and \emph{``European''}. In fact, Topic 2540 (Red) seems to be completely about the automobiles manufacturer Honda, regardless of nature or connotation of the main event. For example, both arguments \emph{``Honda HR-V is equipped with many functions''} and \emph{``Honda recalled its Vezel SUV last year''} fall under Topic 2540. In conclusion, it is important to cluster arguments based on words referring to the event, sentiment, or topic rather than the named-entities.

\end{document}